\documentclass[letterpaper, conference]{ieeeconf}
\IEEEoverridecommandlockouts

\usepackage{amsmath,amsfonts}
\usepackage{array}
\usepackage[caption=false,font=normalsize,labelfont=sf,textfont=sf]{subfig}
\usepackage{textcomp}
\usepackage{stfloats}
\usepackage{url}
\usepackage{graphicx}
\usepackage{cite}

\usepackage[draft,bookmarks=false]{hyperref}
\usepackage[flushleft]{threeparttable}
\usepackage{booktabs}
\usepackage{xpatch}
\usepackage{balance}

\graphicspath{{./imgs/}}

\title{\LARGE \bf PSR: Predictive Sensorimotor Representation Learning for \\ Contact-Rich Manipulation}

\author{Shengbao Li$^{*}$, Peng Xu$^{*}$, Chao Tang, Hao Wei, Jiaheng Wang, Hong Yin,\\
Jiangtao Chen, Jinxuan Zhu, Zhong Zhou, Mengfan Wang, Tingguang Li$^{\dagger}$\\
Samsung R\&D Institute China-Beijing\thanks{$^{*}$Equal contribution. $^{\dagger}$Corresponding author.}}

\makeatletter
\xpatchcmd{\@maketitle}{\vskip0.25in}{\vskip0.05in}{}{\PackageError{title-spacing}{Top spacing patch failed}{}}
\xpatchcmd{\@maketitle}{\vskip1.0em\par}{\vskip0.4em\par}{}{\PackageError{title-spacing}{Author spacing patch failed}{}}
\makeatother
\IEEEaftertitletext{\vspace{-\baselineskip}}

\begin{document}

\maketitle
\thispagestyle{empty}
\pagestyle{empty}

\suppressfloats[t]
\begin{abstract}

Contact-rich manipulation requires policies to generate precise actions by reasoning over contact forces, robot configurations, and interaction histories beyond visual observations. Existing methods passively condition on force feedback rather than actively predicting future contact dynamics, limiting their ability to generate high-precision actions. To address this problem, we introduce Predictive Sensorimotor Representation (PSR) learning, a framework that learns a hierarchy of predictive representations from multimodal sensorimotor signals and integrates them into the action stream of a visuomotor policy. Specifically, during a pretraining stage, a multimodal Transformer is trained to learn a hierarchy of predictive representations by jointly forecasting future interaction dynamics. The learned hierarchy subsequently augments the action stream, enabling the resulting policy to exploit contact-relevant cues at multiple depths. We further instantiate PSR within a Vision-Language-Action (VLA) model, resulting in \textbf{PSR-VLA}, and evaluate it on six real-world contact-rich manipulation tasks. Experimental results show that PSR-VLA achieves $91.7\%$ overall success, improving over $\pi_{0.5}$, ForceVLA-$\pi_{0.5}$, and ForceVLA2-$\pi_{0.5}$ by $30.0$, $22.5$, and $19.2$ percentage points, respectively. These results demonstrate the effectiveness of the proposed PSR for force-aware, contact-rich manipulation. Videos of the tasks and stability tests are available at
{\urlstyle{same}\mbox{\nolinkurl{https://psr-vla.pages.dev/}}}.

\end{abstract}

\section{Introduction}

Contact-rich manipulation requires a robot to reason about how physical interactions evolve over time beyond visual observations. In complex tasks such as insertion, wiping, and assembly, visually similar states may correspond to different actions due to variations in current and historical contact states. Misinterpreting this information can lead to jamming, slipping, excessive force, or even failed execution.

\begin{figure}[t]
\centering
\includegraphics[width=\columnwidth]{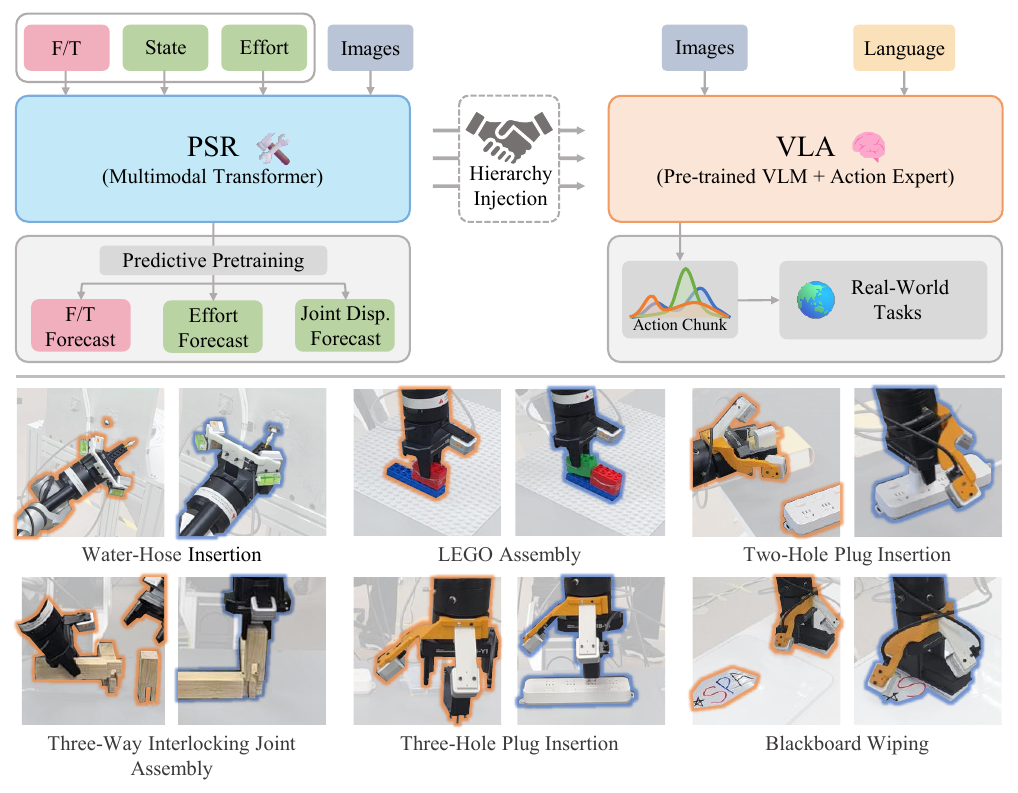}
\vspace{-0.3in}
\caption{An overview of PSR-VLA and the six real-world contact-rich manipulation tasks. PSR first learns a hierarchy of predictive representations from multimodal sensorimotor signals and integrates them into the action stream of a visuomotor policy for improved precision and robustness.}
\label{fig:intro_overview}
\vspace{-0.3in}
\end{figure}

Although Vision-Language-Action (VLA) models \cite{brohan2023rt1,zitkovich2023rt2,kim2025openvla,black2025pi0,black2025pi05} inherit broad semantic understanding capabilities from large pre-trained VLMs, they remain limited in modeling the dynamics of contact-rich interaction. Recent approaches address this limitation by incorporating force and torque feedback as additional sensory inputs \cite{yu2025forcevla,li2026forcevla2,li2026atvla}. However, they primarily use these signals as reactive conditioning inputs rather than predictive learning targets, and thus do not explicitly anticipate how contact dynamics will evolve. This limits their ability to generate precise actions, particularly during rapidly changing or sustained contact interactions.

In this paper, we argue that actively predicting future sensorimotor signals provides an effective mechanism for modeling evolving contact dynamics. Following this idea, we propose \textbf{Predictive Sensorimotor Representation (PSR) learning}, a framework that learns a hierarchy of predictive representations from multimodal sensorimotor signals and can be seamlessly integrated into the action stream of a visuomotor policy for contact-rich manipulation. Technically, PSR operates in two stages. During Stage~1 of predictive pretraining, PSR learns a hierarchy of predictive sensorimotor representations by jointly forecasting future force/torque sequences, joint-effort sequences, and terminal joint-state displacement with a multimodal Transformer. More specifically, the force/torque and joint-effort targets capture forthcoming contact responses and actuation loads, while terminal joint-state displacement encourages the representations to encode task-level motion intent. During Stage~2 of action learning, PSR augments the policy by integrating the resulting representations into its action stream via gated cross-attention, influencing action generation across multiple depths. We further instantiate PSR within $\pi_{0.5}$, yielding PSR-VLA, and evaluate it on six real-world, contact-rich manipulation tasks spanning insertion, wiping, and assembly. Overall, PSR-VLA achieves an average success rate of 91.7\% across six tasks, outperforming $\pi_{0.5}$, ForceVLA-$\pi_{0.5}$, and ForceVLA2-$\pi_{0.5}$ by 30.0, 22.5, and 19.2 percentage points, respectively. These results demonstrate the effectiveness of the proposed PSR for force-aware, contact-rich manipulation.

\textbf{Contribution}: We propose a novel framework that performs predictive modeling of contact dynamics for force-aware, contact-rich manipulation. It demonstrates improved performance compared to the previous force-aware methods.

\section{Related Work}
\label{sec:related}

\subsection{General VLA}
\label{sec:general_vla}

Vision-language-action (VLA) models transfer generalizable vision-language priors from vision-language models to robot control~\cite{brohan2023rt1,zitkovich2023rt2,kim2025openvla,black2025pi0,black2025pi05}. By grounding language instructions in visual observations and decoding continuous actions, they provide a general interface for manipulation. However, visual-language observations alone contain limited information about physical interaction. Visually similar states can correspond to different contact forces, joint loads, and interaction progress, especially in contact-rich tasks. This makes physical feedback, particularly force/torque and joint-effort histories, important for reliable manipulation and motivates the force-aware methods reviewed next.

\subsection{Force-Aware Manipulation}

Existing force-aware approaches can generally be organized by how they use these signals. Perceptual-conditioning methods use force or torque measurements as policy inputs \cite{yu2025forcevla,hao2026tla,li2026atvla}, whereas auxiliary-prediction methods introduce force-related targets to support action learning \cite{zhang2025torquevla,li2026favla}. ForceVLA2 \cite{li2026forcevla2} uses force both as a policy input and as an explicit control target to generate hybrid force-position commands and estimate subtask progress. These methods use physical signals directly for action learning or control. In contrast, PSR forecasts future interaction dynamics as a representation-learning objective, providing richer predictive context for high-precision action generation.

\subsection{Predictive Representation Learning}

Predictive representation learning captures dependencies in robot experience by reconstructing missing signals or forecasting future ones. RPT \cite{radosavovic2023sensorimotor} first proposes to recover masked content from sequences of images, proprioceptive states, and actions. Later in VLAs, predictive objectives have largely focused on visual dynamics: CoT-VLA \cite{zhao2025cotvla} generates future visual states as reasoning traces, while JEPA-VLA learns video-predictive embeddings that condition multiple policy layers \cite{miao2026jepavla}. For contact-rich manipulation, early visual-tactile methods learn joint representations of vision and touch \cite{lee2020visiontouch}. Building on this foundation, DreamTacVLA \cite{ye2025dreamtacvla} and UniTacVLA \cite{zhang2026unitacvla} forecast tactile futures to guide action generation. A parallel line extends world models to physical interaction: FAWAM \cite{he2026fawam} models force-conditioned dynamics, while Dream-Tac \cite{lou2026dreamtac} models tactile-conditioned dynamics. PSR instead couples external contact response, internal actuation load, and resulting state change within a shared predictive representation, providing broader physical context for high-precision action generation.

\section{Method}
\label{sec:method}

In this section, we first formulate the problem in Section~\ref{sec:problem_formulation}. We then introduce the predictive pretraining objective for Stage~1 in Section~\ref{sec:psr_pretraining} and the multimodal predictive Transformer used to realize this objective in Section~\ref{sec:psr_transformer}. Finally, Section~\ref{sec:psr_vla} describes Stage~2, in which the learned representations are incorporated into the policy’s action stream at multiple depths.

\begin{figure*}[t]
\centering
\includegraphics[width=\textwidth]{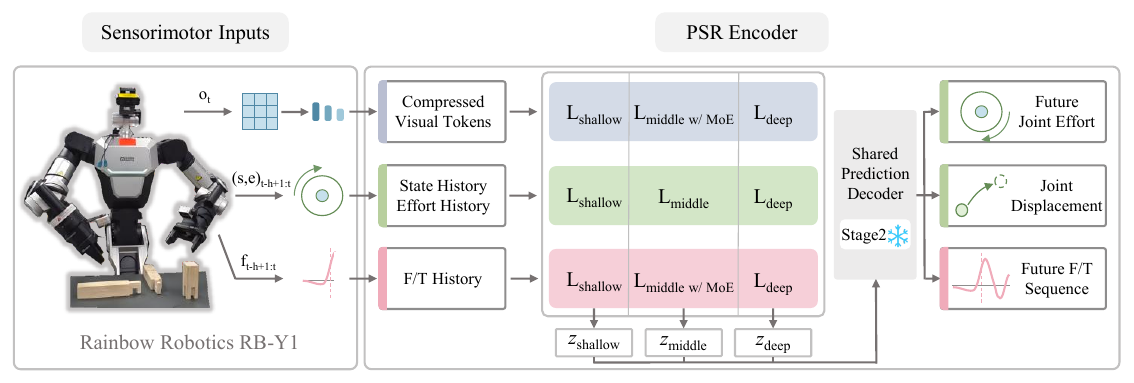}
\caption{PSR learns a hierarchy of predictive sensorimotor representations through multi-target prediction in Stage~1. In Stage~2, PSR integrates the resulting representations into the action stream of a policy via gated cross-attention, influencing action generation across multiple depths.}
\label{fig:method_overview}
\end{figure*}

\subsection{Problem Formulation}
\label{sec:problem_formulation}

We consider language-conditioned, contact-rich manipulation from multimodal inputs. At step \(t\), the robot observes a history of \(h\) sensorimotor frames:
\[
\mathcal{H}_t =
\left(
\mathbf{f}_{t-h+1:t},
\mathbf{s}_{t-h+1:t},
\mathbf{e}_{t-h+1:t}
\right),
\]
where \(\mathbf{f}\), \(\mathbf{s}\), and \(\mathbf{e}\) denote F/T measurements, joint states, and joint efforts, respectively. Given a language instruction $l$, multi-view RGB images $o_t$, and \(\mathcal{H}_t\), the policy \(\pi_\theta\) predicts a continuous \(k\)-step action chunk:
\[
{\mathbf{a}}_{t:t+k-1}
=
\pi_\theta\!\left(
l,o_t,\mathcal{H}_t
\right).
\]

\subsection{Predictive Pretraining}
\label{sec:psr_pretraining}

In the first stage (Fig.~\ref{fig:method_overview}), PSR learns a hierarchy of predictive sensorimotor representations from visual, F/T, and proprioceptive inputs. Visual tokens are reused from the policy backbone and compressed into a fixed number of slots. F/T histories form F/T tokens, while temporally aligned joint states and efforts form proprioceptive tokens.

Contact-rich action generation may require fine-grained force transients, effort changes, and broader multimodal context at different stages. PSR uses a six-layer predictive Transformer and retains every layer output as \(\mathcal{Z}=[z_1,\ldots,z_6]\), a sensorimotor hierarchy indexed by encoder depth. Shallow layers process modalities independently, intermediate layers introduce bottleneck-mediated cross-modal exchange, and the final layer refines the fused tokens. The representations differ in processing depth and multimodal context, providing an ordered hierarchy for policy integration.

The predictive objective comprises future F/T sequences, future joint-effort
sequences, and terminal joint-state displacement, which describe contact
responses, actuation loads, and resulting robot motion, respectively.
This pretraining is action-unconditioned: all three targets are predicted from
observed visual and sensorimotor history, without future action chunks as inputs.
The terminal target is \(\Delta\mathbf{s}_t=\mathbf{s}_{t+p}-\mathbf{s}_t\),
where \(p=k\). It provides coarse-grained motion supervision at the action
horizon without reconstructing intermediate actions or joint states; multiple
trajectories can share the same endpoint displacement.
Using learned level embeddings, target-specific queries in a shared decoder
cross-attend to all depths of \(\mathcal{Z}\), and separate output heads produce
the three predictions. The shared decoder jointly supervises the complete sensorimotor hierarchy;
no separate prediction objective is assigned to each \(z_i\). The predictive pretraining loss is defined as:
\begin{equation}
\label{eq:psr_objective}
\mathcal{L}_{\mathrm{pred}}
=
\alpha\mathcal{L}_{\mathrm{ft}}
+
\beta\mathcal{L}_{\mathrm{eff}}
+
\gamma\mathcal{L}_{\mathrm{disp}}
+
\delta\mathcal{L}_{\mathrm{route}}.
\end{equation}
Here, \(\mathcal{L}_{\mathrm{ft}}\), \(\mathcal{L}_{\mathrm{eff}}\), and
\(\mathcal{L}_{\mathrm{disp}}\) denote mean squared error (MSE) losses for future F/T
sequences, future joint-effort sequences, and terminal joint-state
displacement, respectively. The routing loss \(\mathcal{L}_{\mathrm{route}}\)  and internal architecture used to construct this hierarchy are detailed in the next subsection.

\subsection{Multimodal Predictive Transformer}
\label{sec:psr_transformer}

To construct the sensorimotor hierarchy $\mathcal{Z}$ introduced above, PSR organizes its multimodal Transformer into shallow, intermediate, and deep layers. At shallow depths, the visual, F/T, and proprioceptive streams are processed independently to preserve modality-specific cues before fusion.  At intermediate depths, PSR replaces the dense feed-forward sublayers in the visual and F/T streams with sparsely gated mixture-of-experts (MoE) layers, each comprising four experts and using Top-\(2\) routing~\cite{lepikhin2021gshard}, allowing specialized experts to capture heterogeneous visual and contact patterns. The simpler proprioceptive stream retains a standard dense feed-forward layer to preserve stable robot-state information.

Inspired by attention bottlenecks for multimodal fusion~\cite{nagrani2021attention}, PSR mediates cross-modal exchange at intermediate depths through learned bottleneck tokens, which provide a fixed-capacity workspace for selectively aggregating multimodal information and returning relevant context to each modality. Let \(v_i\), \(f_i\), and \(p_i\) denote the visual, F/T, and proprioceptive tokens at depth \(i\), respectively, and let \(x_i=[v_i;f_i;p_i]\) denote their concatenation. Let \(b_i\) denote the learned bottleneck tokens. \begingroup
\postdisplaypenalty=10000
Equation~\eqref{eq:psr_fusion} applies only to intermediate fusion layers:
\begin{equation}
\begin{aligned}
\widetilde{b}_i &= b_i +
\operatorname{CA}_{b\leftarrow x}\!\left(b_i,x_i,x_i\right),\\
z_i &= x_i +
g_i\,\operatorname{CA}_{x\leftarrow b}\!\left(x_i,\widetilde{b}_i,\widetilde{b}_i\right).
\end{aligned}
\label{eq:psr_fusion}
\end{equation}
where \(\operatorname{CA}_{b\leftarrow x}\) and \(\operatorname{CA}_{x\leftarrow b}\) denote cross-attention from the modality tokens to the bottleneck tokens and from the bottleneck tokens to the modality tokens, respectively. The first operation aggregates multimodal cues into \(\widetilde{b}_i\), while the second uses the updated bottleneck tokens to return the aggregated context to the modality tokens. The learnable sigmoid gate \(g_i\) controls the contribution of this feedback. Together with the residual connection to \(x_i\), this bidirectional bottleneck enables controlled cross-modal exchange while preserving modality-specific
information.
\par\endgroup

Because cross-modal exchange begins at intermediate depths and is refined thereafter, the resulting \(z_i\) differ in processing depth and available multimodal context. The routing regularizer \(\mathcal{L}_{\mathrm{route}}\) equally weights the load-balancing penalty and router z-loss~\cite{zoph2022stmoe}, averages their sum over all MoE layers, and is weighted by \(\delta\) during pretraining.

\subsection{Multi-Depth Action Augmentation}
\label{sec:psr_vla}
\begin{figure}[!b]
\centering
\includegraphics[width=0.74\columnwidth,trim=3.5bp 3.5bp 3.5bp 3.5bp,clip]{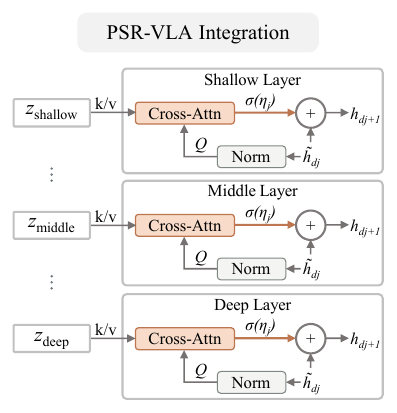}
\caption{Multi-depth action augmentation through gated cross-attention.}
\vspace{-0.1in}
\label{fig:psr_vla_integration}
\end{figure}

To incorporate the hierarchy learned during predictive pretraining into action
generation, we inject PSR representations into selected Action Expert layers at multipl depths (Fig.~\ref{fig:psr_vla_integration}).
Let \(z_j\) denote a PSR representation and \(d_j\) the corresponding depth-aligned Action Expert layer. With vision-language context \(c_{d_j}\) and
incoming action tokens \(h_{d_j}\), the update is
\begin{equation}
\begin{gathered}
\widetilde{h}_{d_j}=\operatorname{AE}_{d_j}(h_{d_j},c_{d_j}),
\quad q_j=\operatorname{Norm}(\widetilde{h}_{d_j}),\\
h_{d_j+1}=\widetilde{h}_{d_j}
+\sigma(\eta_j)\,\operatorname{CA}^{(j)}_{h\leftarrow z}(q_j,z_j,z_j).
\end{gathered}
\label{eq:psr_joint_interaction}
\end{equation}
Here, \(\operatorname{AE}_{d_j}\) is the original Action Expert layer and
\(\widetilde{h}_{d_j}\) is its vision-language-conditioned action state.
In the \(j\)-th injection module, its normalized state \(q_j\) provides the
cross-attention queries and \(z_j\) provides the keys and values. The
learnable sigmoid gate \(\sigma(\eta_j)\) controls the residual sensorimotor
update. Other layers retain their original computation.

During Stage~1, the entire VLA, including the vision-language backbone and Action Expert, remains frozen; only the PSR encoder and the training-only prediction decoder are optimized through sensorimotor prediction. In Stage~2, we jointly fine-tune all parameters of the vision-language backbone, Action Expert, PSR encoder, and gated cross-attention modules using only the policy's action-generation objective. The prediction decoder remains frozen and is not used for action generation; no prediction loss is applied. Predictive pretraining therefore provides the initialization for
action learning rather than a persistent forecasting constraint on the final
policy. At each control step, \(\mathcal{Z}\) is computed once from the current
history and reused throughout action generation. We instantiate the policy with $\pi_{0.5}$ and use its original flow-matching objective.

\begin{table*}[t]
\centering
\begin{threeparttable}
\caption{Main results on six real-world tasks.}
\label{tab:main_results}
\footnotesize
\setlength{\tabcolsep}{3.8pt}
\renewcommand{\arraystretch}{1.08}
\begin{tabular}{@{}l@{\hspace{8pt}}cc@{\hspace{11pt}}cc@{\hspace{11pt}}cc@{\hspace{8pt}}c@{}}
\toprule
\textbf{Method}
& \multicolumn{2}{c}{\textbf{Hard}}
& \multicolumn{2}{c}{\textbf{Medium}}
& \multicolumn{2}{c}{\textbf{Easy}}
& \textbf{Overall} \\
\cmidrule(lr){2-3}
\cmidrule(lr){4-5}
\cmidrule(lr){6-7}
& \shortstack{Three-Way Interlocking\\Joint Assembly}
& \shortstack{Water-Hose\\Insertion}
& \shortstack{Three-Hole Plug\\Insertion}
& \shortstack{LEGO\\Assembly}
& \shortstack{Two-Hole Plug\\Insertion}
& \shortstack{Blackboard\\Wiping}
& \\
\midrule
$\pi_{0.5}$
& 9/20 & 13/20 & 12/20 & 13/20 & 13/20 & 14/20 & 74/120 (61.7\%) \\
ForceVLA-$\pi_{0.5}$
& 12/20 & 14/20 & 12/20 & 14/20 & 15/20 & 16/20 & 83/120 (69.2\%) \\
ForceVLA2-$\pi_{0.5}$
& 13/20 & 15/20 & \textbf{17/20} & 11/20 & 16/20 & 15/20 & 87/120 (72.5\%) \\
\midrule
\textbf{PSR-VLA}
& \textbf{18/20} & \textbf{19/20} & 16/20 & \textbf{18/20} & \textbf{19/20} & \textbf{20/20} & \textbf{110/120 (91.7\%)} \\
\bottomrule
\end{tabular}
\begin{tablenotes}[flushleft]
\footnotesize\normalfont
\item[] \textit{Notes.} Each method uses 20 trials per task. Overall reports total successes over 120 trials, with success rates in parentheses. ForceVLA2-$\pi_{0.5}$ uses its short-horizon force-conditioning pathway adapted to the common action-generation protocol.
\end{tablenotes}
\end{threeparttable}
\vspace{-0.8em}
\end{table*}

\section{Experimental Setup}
\label{sec:exp_setup}

\textbf{Baselines.}
We compare sensorimotor conditioning designs under a common action-generation protocol. All methods use the same $\pi_{0.5}$ backbone, demonstrations, action representation, and action horizon. The original $\pi_{0.5}$ uses no F/T input. For ForceVLA-$\pi_{0.5}$ \cite{yu2025forcevla}, we replace only its original $\pi_0$ backbone while retaining its complete force-aware expert-routing mechanism. For ForceVLA2-$\pi_{0.5}$ \cite{li2026forcevla2}, we adapt only its short-horizon force-conditioning pathway to the common action representation and horizon, rather than its full hybrid force-position controller. We retain its cross-modal physical-state interaction, direct F/T bypass, and token-wise Top-1 three-expert Cross-Scale MoE; force prompting and progress-transition supervision are excluded because they require method-specific prompts, annotations, and control variables. This baseline evaluates the adapted sensor-conditioning pathway under that protocol. Router diagnostics over complete forward passes confirmed that all three experts remained active, with no collapse to a single expert. PSR-VLA's method-specific predictive pretraining reuses these demonstrations and introduces no additional training data. Sensor inputs, parameter counts, and pretraining budgets differ across methods.

\textbf{Real-World Tasks.}
We evaluate all methods on six contact-rich manipulation tasks, grouped as hard
(Three-Way Interlocking Joint Assembly and Water-Hose Insertion), medium (LEGO
Assembly and Three-Hole Plug Insertion), and easy (Two-Hole Plug Insertion and
Blackboard Wiping). Collectively, these tasks span rigid and deformable
insertion, multi-part assembly, precise multi-contact alignment, and sustained
force-regulated contact.

The Rainbow Robotics RB-Y1 uses a head-mounted Stereolabs ZED X Mini, two wrist-mounted Intel RealSense D405 cameras (synchronized), bilateral six-axis wrist F/T sensors, and joint-state and joint-effort measurements. F/T sensors are zero-bias calibrated before use without gravity compensation.

\textbf{Evaluation Protocol.}
For each method, we train a separate task-specific model for each of the six
tasks. Table~\ref{tab:main_results} reports one training run for each
method--task pair. We conduct $20$ consecutive trials per method--task pair
under a task-specific variation schedule fixed before evaluation. For LEGO
Assembly, the first $10$ trials vary position while keeping the LEGO component
in a fixed horizontal orientation, and the remaining $10$ vary both position
and rotation. For Three-Way Interlocking Joint Assembly, the first $10$ trials
vary position only with orientation fixed, and the remaining $10$ vary both
position and rotation. For Water-Hose Insertion, the first $10$ trials use a
fixed initial hose position, and the remaining $10$ randomize its initial
position. For both Three-Hole Plug Insertion and Two-Hole Plug Insertion, the
plug's initial position is independently randomized in all $20$ trials.
Blackboard Wiping comprises six color-controlled trials, two each for blue,
red, and black, plus $14$ trials with two marks placed at random locations.
The robot initial state is independently randomized in every trial; object-pose
variation follows the schedule above and the same task-specific distributions
for all methods. All attempted trials are included, with no retries or post-hoc
exclusions, resulting in $120$ real-robot trials per method.
All offline evaluation episodes use the same experimental settings but are
excluded from all training data.

A trial is successful only if it satisfies the predefined task-specific
completion criterion, and no partial credit is awarded: LEGO Assembly and
Three-Way Interlocking Joint Assembly require every stage to succeed;
Water-Hose Insertion requires full insertion to the endpoint; both plug tasks
require a fully seated, stable fit without looseness or partial insertion; and
Blackboard Wiping requires every mark to be completely removed. We report the
success rate for each task.

\textbf{Implementation Details.}
The PSR encoder has six layers: two modality-independent, three fusion, and one refinement. Outputs $z_1,\ldots,z_6$ are injected, respectively, into layers $3,6,9,12,15,18$ of the 18-layer Action Expert. We use $h=12$ history frames ($0.4\,\mathrm{s}$ at $30\,\mathrm{Hz}$), matching F/T, joint-effort, and action horizons $p=k=30$ ($1.0\,\mathrm{s}$), and $(\alpha,\beta,\gamma,\delta)=(1,1,0.1,0.001)$ for the losses in Eq.~\eqref{eq:psr_objective}. F/T, joint-effort, and displacement channels use per-channel quantile normalization using $q_{01},q_{99}$ from each task's training set, shared across histories, targets, and held-out evaluation. Stage~1 errors use this normalized space and exclude zero-padded displacement dimensions via the training-time mask.

The update/freeze scheme follows Section~\ref{sec:psr_vla}. Stage~1 reuses the demonstrations for $10{,}000$ steps. Both stages use AdamW, global batch $32$, and $5{,}000$-step warmup to $2\times10^{-5}$ followed by a $50{,}000$-step cosine schedule toward $2\times10^{-6}$. Stage~2 uses common demonstrations, action representation/horizon, preprocessing, and optimizer settings across methods, with $30{,}000$ (hard/medium) or $10{,}000$ (easy) action-supervised updates on eight NVIDIA H100 GPUs. The PSR encoder has 1.17B parameters excluding $\pi_{0.5}$; each MoE layer evaluates two of four experts per routed token (Top-$2$). On NVIDIA GeForce RTX 5080, PSR-VLA adds $10\,\mathrm{ms}$ real-robot inference-server latency per request over $\pi_{0.5}$, averaged across all six tasks.

Our $1{,}210$ demonstrations pair the synchronized inputs in Section~\ref{sec:problem_formulation} with action chunks; per-task counts follow Table~\ref{tab:main_results} column order: $(300,300,200,200,150,60)$.

\section{Experimental Results}
\label{sec:exp_results}

\textbf{Main Results.}
Table~\ref{tab:main_results} reports the success rates on six real-world
contact-rich manipulation tasks under the common action-generation protocol in
Section~\ref{sec:exp_setup}. PSR-VLA achieves the highest overall success rate,
with \(110/120\) successful trials (\(91.7\%\)). The tabulated PSR-VLA models use the first completed training run for each task
and were not selected based on real-robot test performance. After completing
the main comparison, we trained three additional sets of PSR-VLA models solely
to assess its training stability. Across the tabulated run and these three
additional runs, PSR-VLA achieves \(109.5 \pm 2.5\) successes out of \(120\)
(\(91.3 \pm 2.1\%\), mean \(\pm\) s.d.). This PSR-VLA-only analysis assesses
whether the tabulated result is representative of its observed training
variation; it is not a seed-matched comparison with the baselines. Under this
protocol, the tabulated PSR-VLA run exceeds $\pi_{0.5}$,
ForceVLA-$\pi_{0.5}$, and an adapted ForceVLA2-$\pi_{0.5}$ short-horizon
conditioning baseline by \(30.0\), \(22.5\), and \(19.2\) percentage points,
respectively.

For $\pi_{0.5}$, ForceVLA-$\pi_{0.5}$, ForceVLA2-$\pi_{0.5}$, and PSR-VLA, descriptive 95\% Wilson intervals for the pooled 120-trial success proportions are 52.7--69.9\%, 60.4--76.7\%, 63.9--79.7\%, and 85.3--95.4\%, respectively. These intervals characterize rollout-level uncertainty for the evaluated models and exclude training-seed variation.

Finally, PSR-VLA achieves the best performance across all three difficulty
groups. On the hard tasks, it reaches an average success rate of \(92.5\%\),
outperforming the strongest baseline by \(22.5\) percentage points. On the
medium and easy tasks, it achieves \(85.0\%\) and \(97.5\%\), with margins of
\(15.0\) and \(20.0\) percentage points, respectively. The largest observed
advantage therefore occurs on the hard tasks, while PSR-VLA remains the
top-performing method on both medium and easy tasks. These results indicate
that the gain is not limited to simpler tasks or near-ceiling settings.

On Three-Hole Plug Insertion, PSR-VLA achieves $16/20$ successes versus $17/20$ for the adapted ForceVLA2-$\pi_{0.5}$ baseline, a one-trial difference that does not establish a systematic performance gap. Under the shared camera setup, limited visibility of the plug--socket relative pose may hinder initial visual alignment.

\begin{figure}[!t]
\centering
\vspace{-0.1in}
\makebox[\columnwidth][c]{\includegraphics[width=\columnwidth]{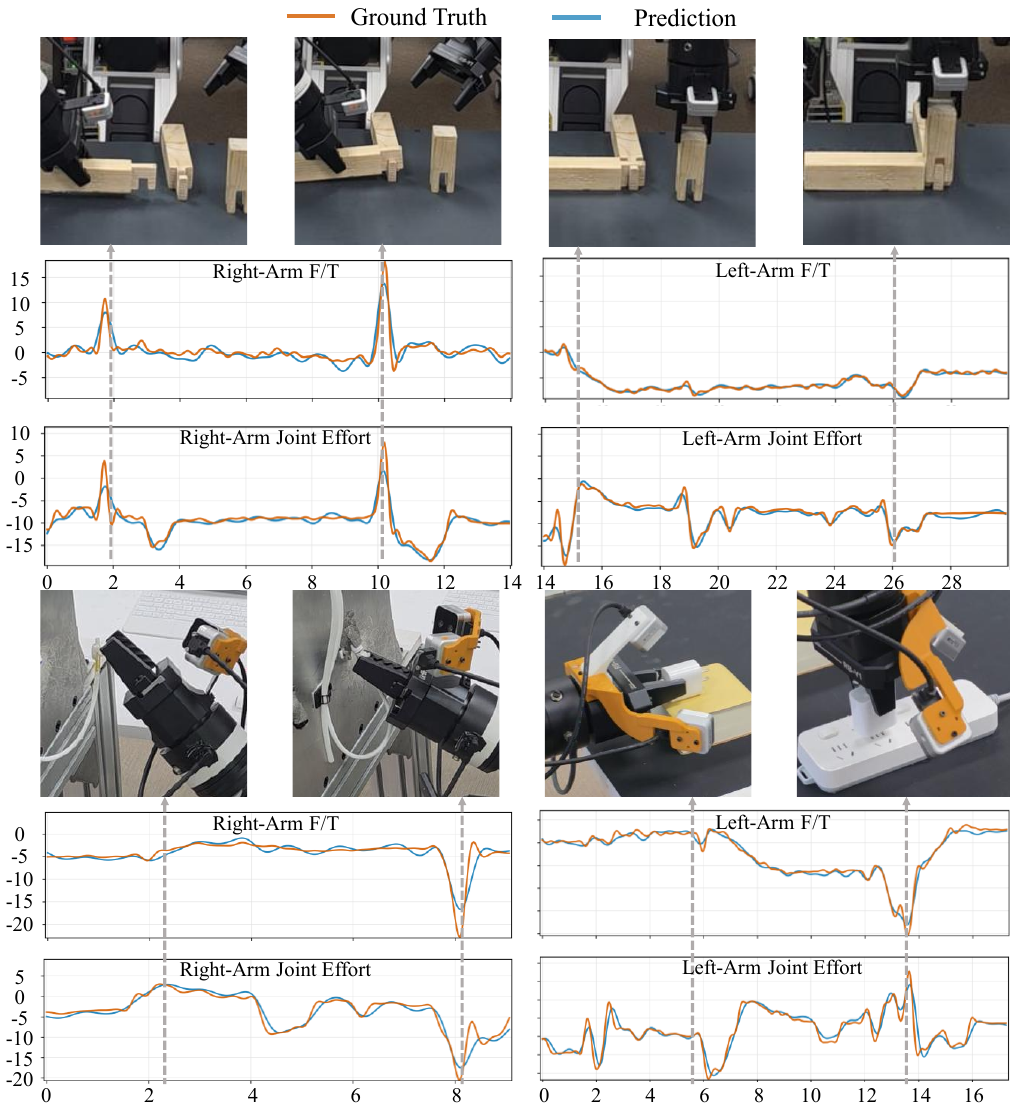}}
\caption{Stage~1 predictions on held-out offline sequences. The decoder predicts all 12 F/T channels, 20 joint-effort channels, and terminal joint-state displacement. The horizontal axes show time (s); the vertical axes show force (N) or joint effort (N\,m), as indicated by the panel titles. Representative curves are selected by event-associated ground-truth variation rather than prediction accuracy; measured and predicted signals are filtered identically.}
\vspace{-0.2in}
\label{fig:experimental_analysis}
\end{figure}

\textbf{Physical Interaction Analysis.}
Fig.~\ref{fig:experimental_analysis} evaluates Stage~1 prediction on 50
held-out offline sequences per task (300 in total).
The displayed traces use physical units (Fig.~\ref{fig:experimental_analysis}), whereas the reported MSEs are computed in normalized space.
Under the training-set quantile normalization described in Section~\ref{sec:exp_setup}, a Persistence baseline repeats the latest observed F/T and joint-effort values over the prediction horizon and predicts zero terminal displacement. It obtains MSEs of \(0.1493\), \(0.0891\), and \(0.9092\) for future F/T sequences, future joint-effort sequences, and terminal joint-state displacement, respectively, whereas PSR obtains \(0.0673\), \(0.0462\), and \(0.1134\). These results correspond to relative reductions of \(54.9\%\), \(48.1\%\), and \(87.5\%\). The displacement metric uses the training-time action mask and excludes zero-padded dimensions. Thus, Stage~1 learns predictive dynamics beyond copying the latest observation, while its downstream control value is assessed separately by the controlled Q1 ablation.

\begin{table}[t]
\centering
\begin{threeparttable}
\caption{Offline action-generation ablations.}
\label{tab:offline_ablation_grouped}
\footnotesize
\setlength{\tabcolsep}{3.0pt}
\renewcommand{\arraystretch}{1.08}
\begin{tabular}{@{}l@{\hspace{5pt}}cc@{\hspace{7pt}}cc@{}}
\toprule
\textbf{Variant}
& \multicolumn{2}{c}{\shortstack{\textbf{Water-Hose}\\\textbf{Insertion}}}
& \multicolumn{2}{c}{\shortstack{\textbf{Two-Hole Plug}\\\textbf{Insertion}}} \\
\cmidrule(lr){2-3}
\cmidrule(lr){4-5}
& \textbf{MSE $\downarrow$} & \textbf{L1 $\downarrow$}
& \textbf{MSE $\downarrow$} & \textbf{L1 $\downarrow$} \\
\midrule
\multicolumn{5}{@{}l}{\textbf{Q1: Predictive Pretraining}} \\
\shortstack[l]{w/o Stage-1\\Pretraining}
& 0.005190 & 0.026432
& 0.004378 & \textbf{0.033470} \\
\textbf{PSR-VLA}
& \textbf{0.003378} & \textbf{0.021413}
& \textbf{0.003816} & 0.035424 \\
\midrule
\multicolumn{5}{@{}l}{\textbf{Q2: Representations and Depth Alignment}} \\
\shortstack[l]{Shared $z_6$\\at All Layers}
& 0.004269 & 0.030268
& 0.004066 & 0.036335 \\
Reversed Alignment
& 0.004696 & 0.029665
& 0.004807 & 0.041436 \\
\textbf{PSR-VLA}
& \textbf{0.003378} & \textbf{0.021413}
& \textbf{0.003816} & \textbf{0.035424} \\
\midrule
\multicolumn{5}{@{}l}{\textbf{Q3: Multi-depth Integration}} \\
\shortstack[l]{Output-Level\\Conditioning}
& 0.005969 & 0.035745
& 0.004438 & 0.041181 \\
\textbf{PSR-VLA}
& \textbf{0.003378} & \textbf{0.021413}
& \textbf{0.003816} & \textbf{0.035424} \\
\bottomrule
\end{tabular}
\begin{tablenotes}[flushleft]
\footnotesize\normalfont
\item[] \textit{Notes.} All variants use the same 100 complete offline episodes per task. MSE and L1 are computed in normalized action space over action-mask-enabled dimensions.
\end{tablenotes}
\end{threeparttable}
\vspace{-0.8em}
\end{table}

\begin{figure}[!t]
\centering
\includegraphics[width=\columnwidth]{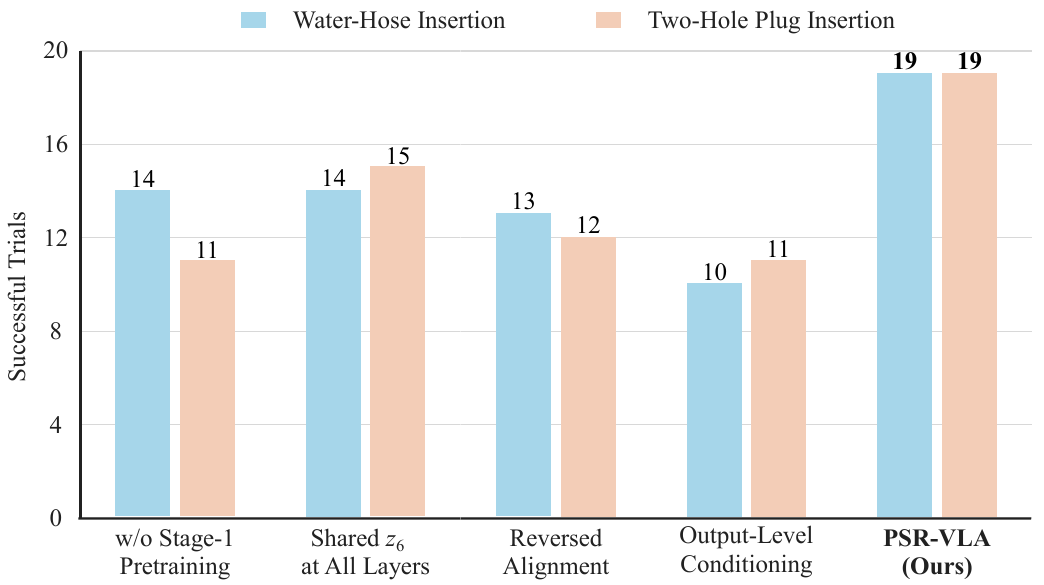}
\caption{Ablation study on Water-Hose Insertion and Two-Hole Plug Insertion. Bars show successes out of $20$ attempts per task.}
\vspace{-0.2in}
\label{fig:ablation}
\end{figure}

\textbf{Ablation Study.}
Before the real-robot trials, all variants are evaluated on the same 100
complete offline episodes per task in normalized action space over
action-mask-enabled dimensions. We report MSE and L1. Together,
Table~\ref{tab:offline_ablation_grouped} and Fig.~\ref{fig:ablation} test
whether (Q1) predictive initialization improves action learning at fixed encoder capacity, (Q2)
distinct representations across encoder depths and their ordered alignment with policy
depth matter, and (Q3) multi-depth integration improves action generation over
Output-Level Conditioning.

These variants assess complementary design choices under the controls specified below. The w/o Stage-1 Pretraining variant retains
the complete PSR-VLA architecture and Stage~2 policy-training budget but
removes predictive initialization. Shared $z_6$ at All Layers supplies
$z_6$ to all six unchanged injection sites. Reversed Alignment retains the
same representations and injection sites but maps $z_i$ to the $(7-i)$-th
selected site---layers $18,15,12,9,6,3$---changing only the pairing order.
Output-Level Conditioning retains Stage~1 but conditions only the Action
Expert output.

\noindent\textbf{Q1: Predictive Pretraining.}\par\nopagebreak[4]

PSR-VLA has lower MSE on both tasks and lower L1 on Water-Hose Insertion,
whereas w/o Stage-1 Pretraining has slightly lower L1 on Two-Hole Plug Insertion; thus, the
offline metrics are not uniformly favorable. The w/o Stage-1 Pretraining variant achieves $70\%$ and $55\%$ real-robot success, compared with $95\%$ on both tasks for PSR-VLA.
Because the architecture and Stage~2 action-supervised budget are unchanged,
this comparison evaluates whether prediction-pretrained initialization improves
action generation beyond the same architecture trained from scratch under an
identical downstream action-training protocol.

\noindent\textbf{Q2: Representations and Depth Alignment.}\par\nopagebreak[4]

These controls provide the most direct evidence for depth-aligned action augmentation.
Reversed Alignment retains all six representation depths, the same injection
sites, and the same number of modules, changing only their pairing order.
Success falls from $95\%$ on each task to $65\%$ on Water-Hose Insertion and
$60\%$ on Two-Hole Plug Insertion, while all four offline errors increase.
Thus, the proposed order outperforms reversed pairing at fixed injection
capacity on these two tasks. Shared $z_6$ at All Layers also
increases all four offline errors and yields $70\%$ and $75\%$ success,
supporting the use of distinct representations across encoder depths at the six injection sites. Together, these results support representation choice and pairing order among the evaluated alternatives without establishing universal optimality.

\noindent\textbf{Q3: Multi-Depth Integration.}

Output-Level Conditioning retains Stage~1 but replaces the six injection
modules with one output-level module. Its higher errors on all four offline
metrics and $50\%$/$55\%$ real-robot success support the evaluated multi-depth
integration design as a whole. This comparison also changes the number of
injection modules, so it does not isolate injection depth from module capacity.

Across the two evaluated tasks, no controlled variant matches PSR-VLA in
real-robot success, and the full model achieves the lowest offline action MSE,
supporting predictive initialization, distinct representations across encoder depths,
multi-depth integration, and the proposed pairing order without implying
universal optimality.

\par\noindent\textbf{PSR-VLA Stress Tests.} We reuse the main-experiment PSR-VLA checkpoints for repeated-trial stress tests without additional data or training; these PSR-VLA-only trials are not matched baseline comparisons. On Water-Hose Insertion, PSR-VLA succeeds in \(442/450\) consecutive trials (\(98.2\%\)) across nine refrigerator positions (50 trials each). On Three-Way Interlocking Joint Assembly, it succeeds in \(48/50\) consecutive trials (\(96.0\%\)). Every attempt is counted, with no recovery or retry. Videos of both evaluations are provided in the supplementary video.

\section{Conclusion}
\label{sec:conclusion}
We present PSR, a framework that learns a hierarchy of predictive sensorimotor representations and uses it to augment the action stream for contact-rich manipulation. During Stage~1 of predictive pretraining, PSR jointly forecasts future F/T sequences, future joint-effort sequences, and terminal joint-state displacement. The learned sensorimotor hierarchy is then integrated into selected Action Expert layers through gated cross-attention at multiple depths. Across six real-world tasks, PSR-VLA achieves a $91.7\%$ overall success rate and outperforms the evaluated VLA baselines under a common action-generation protocol. Ablations on two insertion tasks support predictive pretraining, distinct representations across encoder depths, and their multi-depth integration. These results support learning a sensorimotor hierarchy through future prediction and using it to improve contact-rich action generation. Future work will examine generalization to new objects and contact conditions and extend the framework to manipulation over longer horizons.

\bibliographystyle{IEEEtran}
\balance
\bibliography{root}

\end{document}